\documentclass[
]{ceurart}

\usepackage{listings}
\begin{document}

\copyrightyear{2024}
\copyrightclause{Copyright for this paper by its authors.
  Use permitted under Creative Commons License Attribution 4.0
  International (CC BY 4.0).}

\conference{CLEF 2024: Conference and Labs of the Evaluation Forum, September 09–12, 2024, Grenoble, France}

\title{Overview of PlantCLEF 2024: multi-species plant identification in vegetation plot images}

\tnotemark[1]
\tnotetext[1]{You can use this document as the template for preparing your
  publication. We recommend using the latest version of the ceurart style.}

\title[mode=sub]{Notebook for the LifeCLEF Lab at CLEF 2024}

\author[1,2]{Herv\'e Go\"eau}[%
orcid=0000-0003-3296-3795,
email=herve.goeau@cirad.fr
]
\author[1]{Vincent Espitalier}[%
email=vincent.espitalier@cirad.fr
]
\author[1]{Pierre Bonnet}[%
orcid=0000-0002-2828-4389,
email=pierre.bonnet@cirad.fr
]

\author[2]{Alexis Joly}[%
orcid=0000-0002-2161-9940,
email=alexis.joly@inria.fr
]
\address[1]{CIRAD, UMR AMAP, Montpellier, Occitanie, France}
\address[2]{Inria, LIRMM, Univ Montpellier, CNRS, Montpellier, France}

\cortext[1]{Corresponding author.}
\fntext[1]{These authors contributed equally.}

\begin{abstract}
Plot images are essential for ecological studies, enabling standardized sampling, biodiversity assessment, long-term monitoring and remote, large-scale surveys. Plot images are typically fifty centimetres or one square meter in size, and botanists meticulously identify all the species found there. The integration of AI could significantly improve the efficiency of specialists, helping them to extend the scope and coverage of ecological studies. To evaluate advances in this regard, the PlantCLEF 2024 challenge leverages a new test set of thousands of multi-label images annotated by experts and covering over 800 species. In addition, it provides a large training set of 1.7 million individual plant images as well as state-of-the-art vision transformer models pre-trained on this data. The task is evaluated as a (weakly-labeled) multi-label classification task where the aim is to predict all the plant species present on a high-resolution plot image (using the single-label training data). In this paper, we provide an detailed description of the data, the evaluation methodology, the methods and models employed by the participants and the results achieved.  
\end{abstract}


\maketitle

\section{Introduction}

\begin{figure}
    \centering
    \begin{tabular}{p{0.54\linewidth}|p{0.42\linewidth}}
         \includegraphics[width=\linewidth, trim={14.8cm 0 0cm 0},clip]{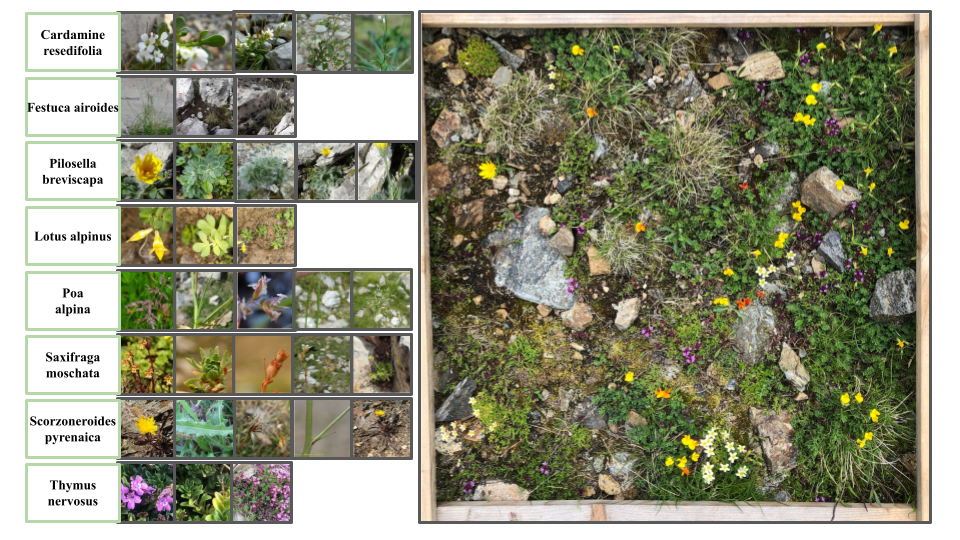} 
         & \includegraphics[width=\linewidth, trim={0 0 19.2cm 0},clip]{Figures/PlantCLEF_2024_one_slide.png} \\ 
         (a) Vegetative plot (test)
         &
         (b) Individual plant images (training)\\
    \end{tabular}    
    \caption{PlantCLEF 2024: illustration of the visual discrepancy between (a) the test set composed exclusively of vertical top views of quadrats with potentially many plant species and (b) the training set based on images of individual plants focusing mostly on organs (flowers, fruits, leaves, stems) with various angles and an wide range of shooting conditions due to the collaborative nature of the original dataset.}
    \label{fig:PlantCLEF2024discrepancy}
\end{figure}

Vegetation plot inventories are essential for ecological studies, as they enable standardized sampling, biodiversity assessment, long-term monitoring and large-scale, remote surveys. They provide valuable data on ecosystems, biodiversity conservation, and evidence-based environmental decision-making. Plot images, also known as "quadrat" are typically 0.5x0.5 meter size, and botanists meticulously identify all the species found there. In addition, they quantify species abundance using indicators such as biomass, qualification factors, and areas occupied in photographs. AI could significantly improve the efficiency of specialists, helping them to extend the frequency and coverage of ecological studies. It could also enable the engagement of non-expert citizen scientists in monitoring programs. Existing plant identification applications (such as Pl@ntNet or iNaturalist) are already a valuable tool in this respect. They can automatically identify specimens present in the plot by photographing them one by one. A much more effective approach, however, would be to be able to directly identify all the individuals in a quadrat by taking a single high-resolution photo of the it (such as the one shown in Figure \ref{fig:PlantCLEF2024discrepancy}). However, the development of AI models solving this multi-label classification task remains challenging. Ideally, we would need very large volumes of quadrat images labeled with all plants present but this would require a colossal amount of expert work to cover the thousands of species of an entire flora. On the other side, very large volumes of labeled images of individual plant specimens are now available and it is easy to train efficient classification models on top of them \cite{plantclef2022,plantclef2023}. Thus, the PlantCLEF2024 challenge proposes to evaluate the task of identifying the species present in quadrat images based on a large training set of individual plant images. As illustrated in Figure \ref{fig:PlantCLEF2024discrepancy}, the main difficulty lies in the shift between both types of data. Whereas the test data is composed of high-resolution multi-label images of vegetation plots (with potentially many species), the training data is composed essentially of single-label images of individual plants or parts of individual plants (coming from the Pl@ntNet collaborative platform \cite{affouard2017pl}). In addition to the different conditions and angles of the shots, the phenological stages can further increase the data disparity. The individual plant data is indeed over represented by close-up views of flowers (facilitating the identification), whereas vegetation plots are most often areas observed multiple times over one or several years, without prior assumptions about the phenological stage of the plants (some may be flowering, others fruiting, some in seedling stage, and others senescent or even affected by a disease). 

\section{Dataset}

\subsection{Training dataset}

The training dataset is composed of observations of individual plants, such as those used in previous editions of PlantCLEF. More precisely, it is a subset of the Pl@ntNet training data focusing on south western Europe and covering 7,806 plant species. It contains about 1.4 million images extended with some images with trusted labels aggregated from the GBIF platform to complete the less illustrated species. Links to original images are provided in the 'url' column of the metadata csv file shared with the participants. The images have a relatively high resolution (minimum or maximum of 800 pixels on the biggest side) to allow the use of classification models that can handle relatively large resolution inputs and may reduce the difficulty of predicting small plants in large vegetative plot images. Images are pre-organized into subfolders by class (i.e., by species) and split into a predefined train-validation-test sets to facilitate the training of individual plant classification models.

\begin{table}
    \caption{PlantCLEF 2024 training dataset statistics. It was provided to participants in the form of three sub-directories, subdividing the data into “Train,” “Val,” and “Test” sets to facilitate the training of individual plant identification models. It is important not to confuse this “Test” set, dedicated to evaluating the performance of potential individual plant identification models, with the challenge test set, which contains large multi-species images.}
    \centering
        \begin{tabular}{cccccc}
            \toprule
            Dataset (with predefined splits) & Images & Observations & Species & Genera & Families \\
            \hline
            All & 1,408,033 & 1,151,904 & 7,806 & 1,446 & 181 \\
            \midrule
            Train & 1,308,899 & 1,052,927 & 7,806 & 1,446 & 181 \\
            Val & 51,194 & 51,045 & 6,670 & 1,415 & 181 \\
            Test & 47,940 & 47,932 & 5,912 & 1,375 & 181 \\            
            \bottomrule
            \end{tabular}
    \label{tab:stats}
\end{table}

\subsection{Test dataset}
 
\begin{figure}
    \centering
    \begin{tabular}{p{0.31\linewidth}p{0.31\linewidth}p{0.31\linewidth}}
        \includegraphics[width=\linewidth,height=\linewidth]{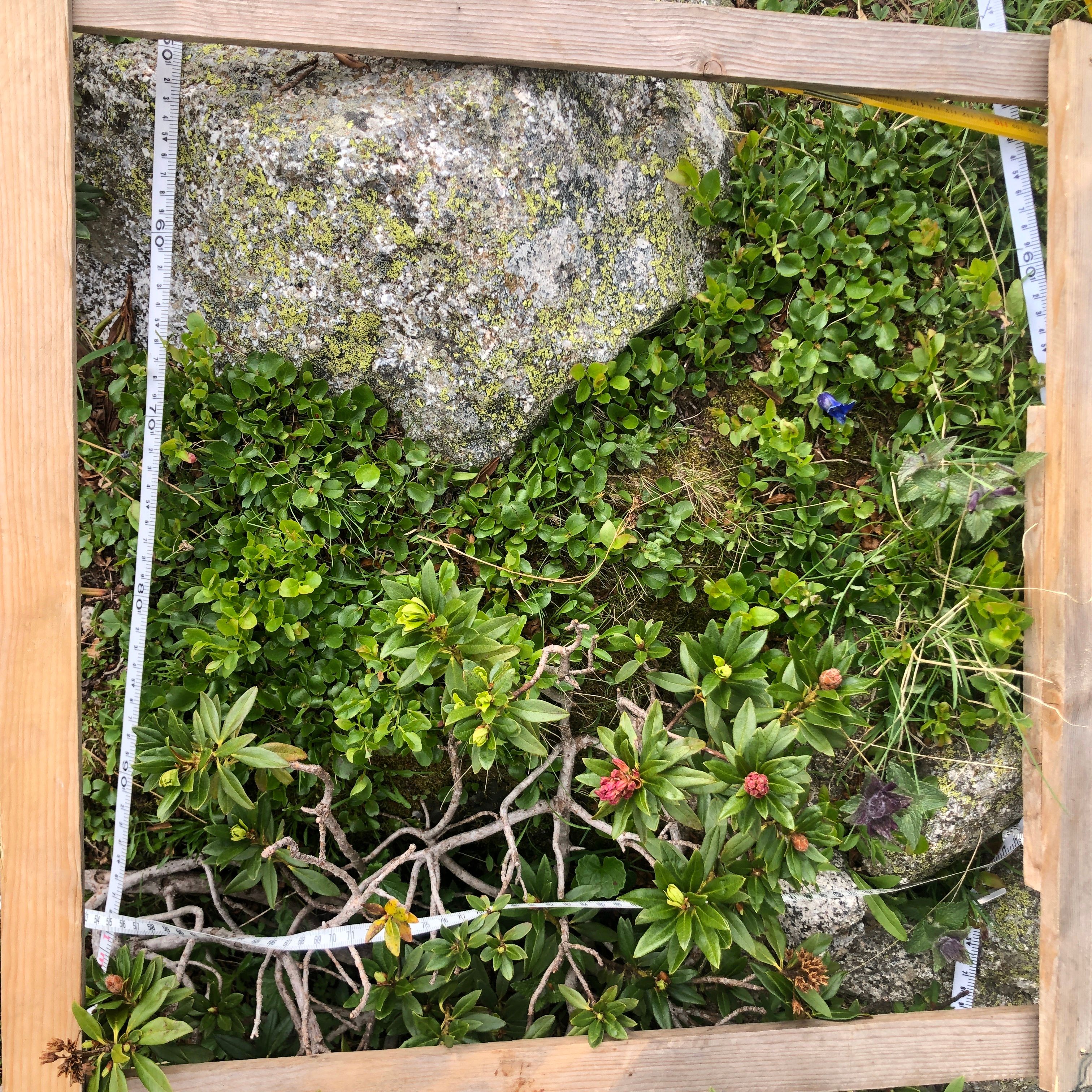} &
        \includegraphics[width=\linewidth,height=\linewidth]{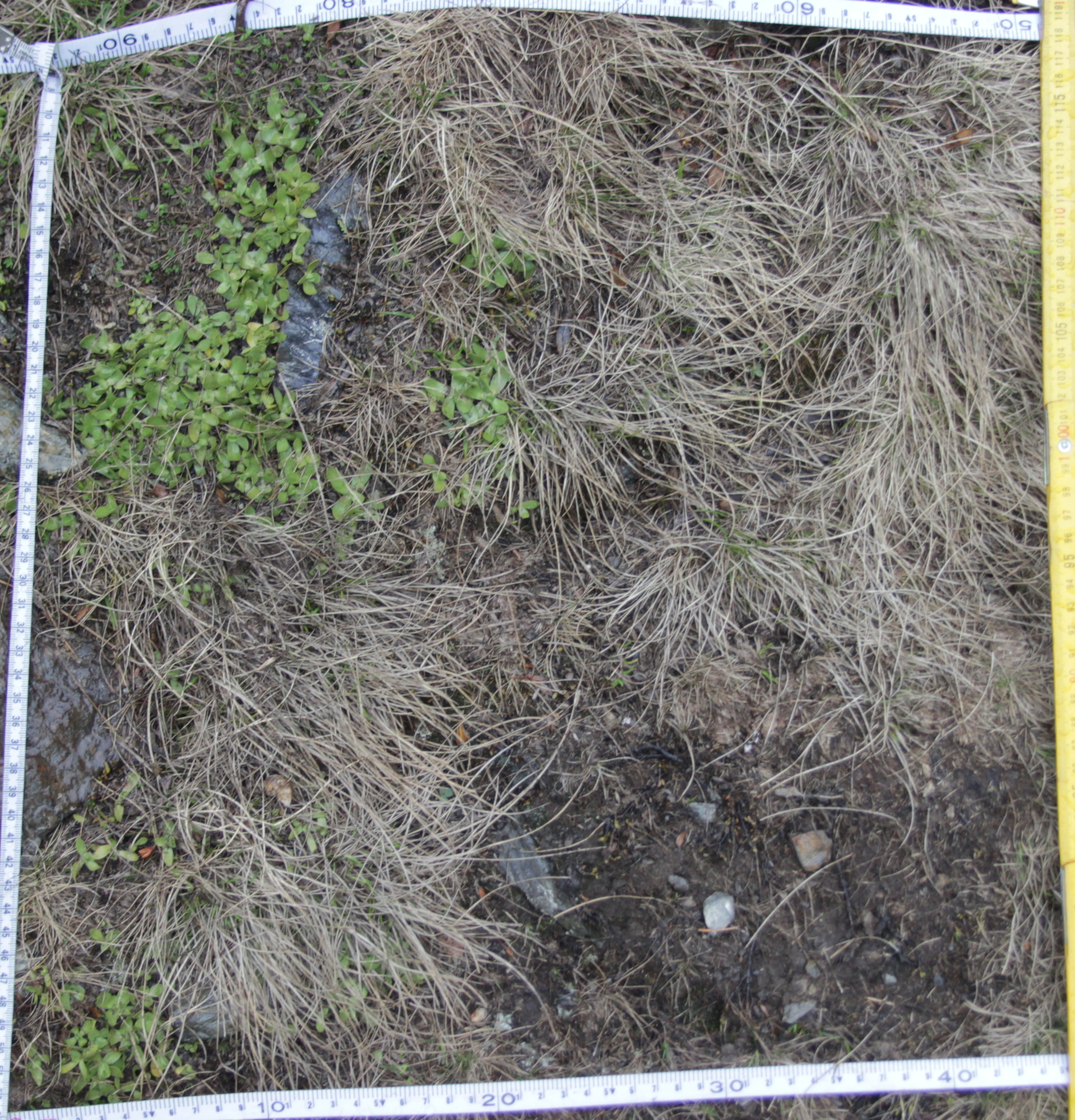} &
        \includegraphics[width=\linewidth,height=\linewidth]{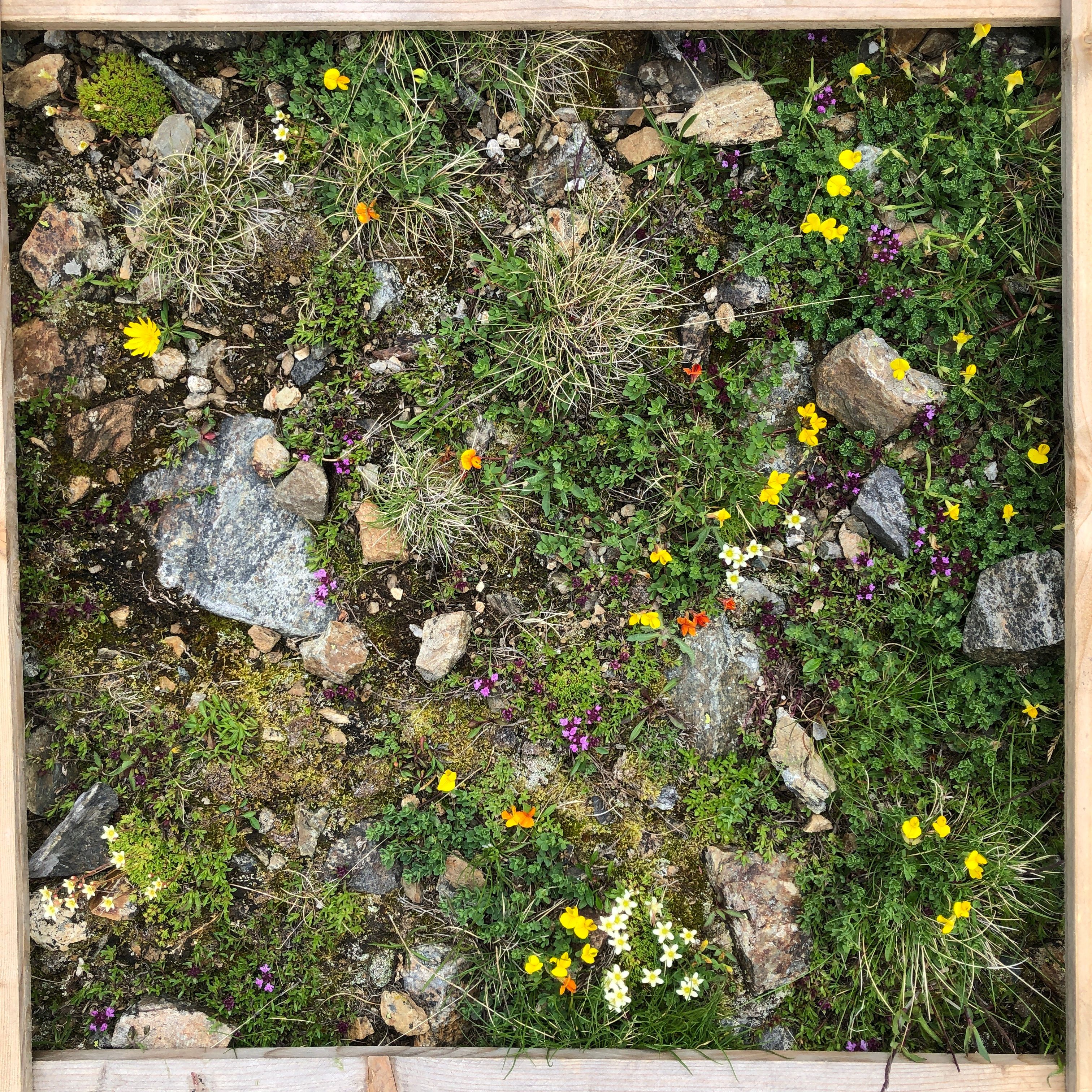} \\
        \includegraphics[width=\linewidth,height=\linewidth]{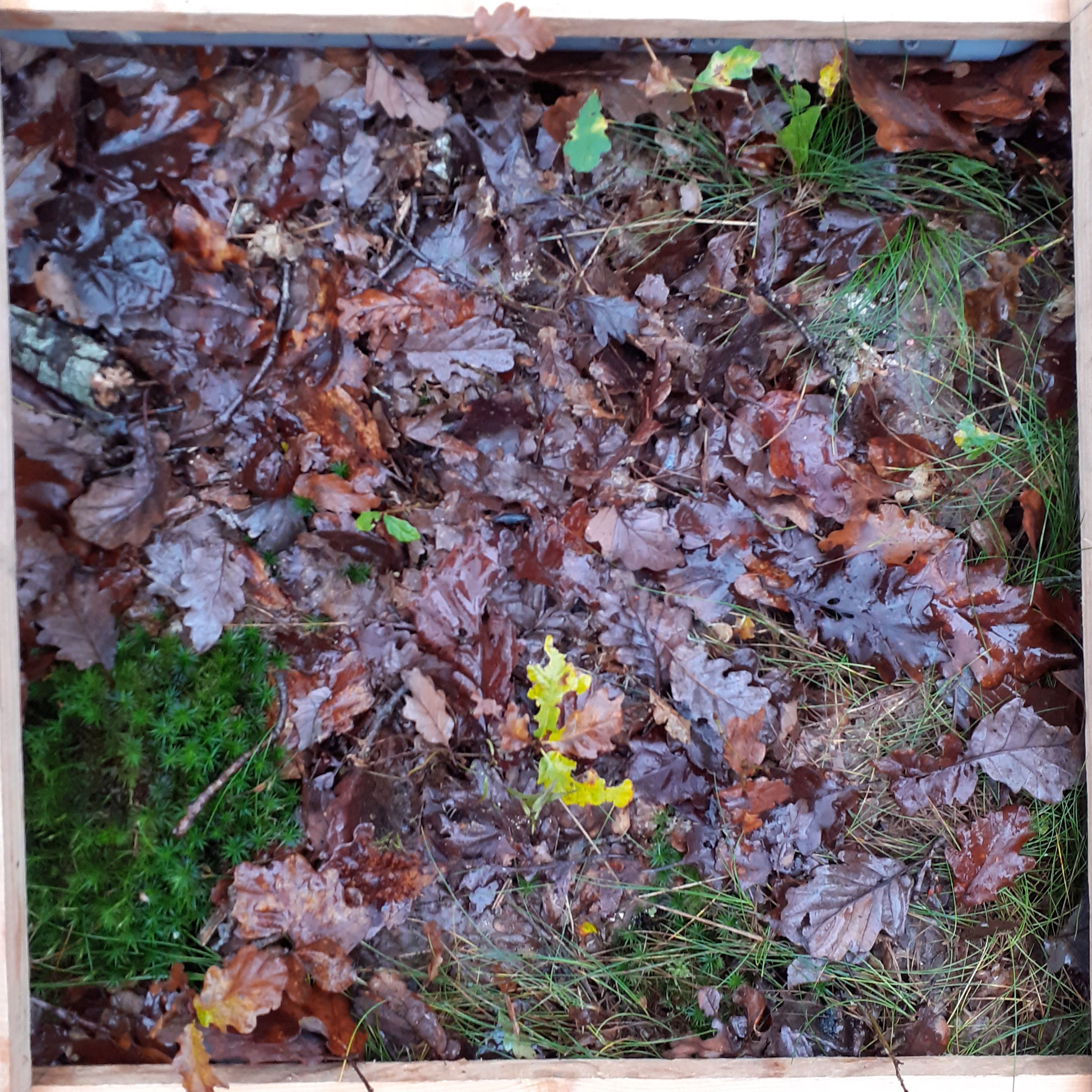} &  
        \includegraphics[width=\linewidth,height=\linewidth]{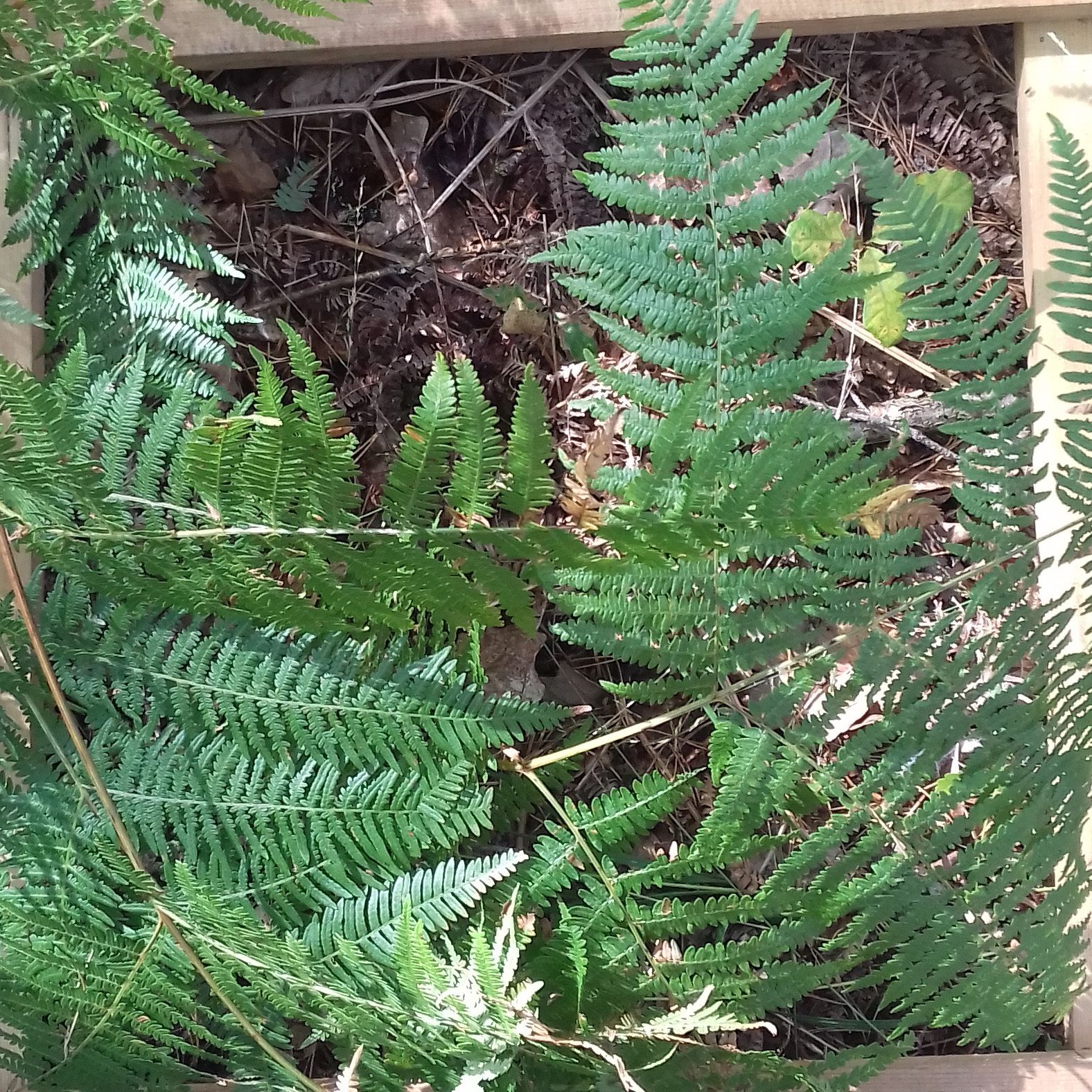} &
        \includegraphics[width=\linewidth,height=\linewidth]{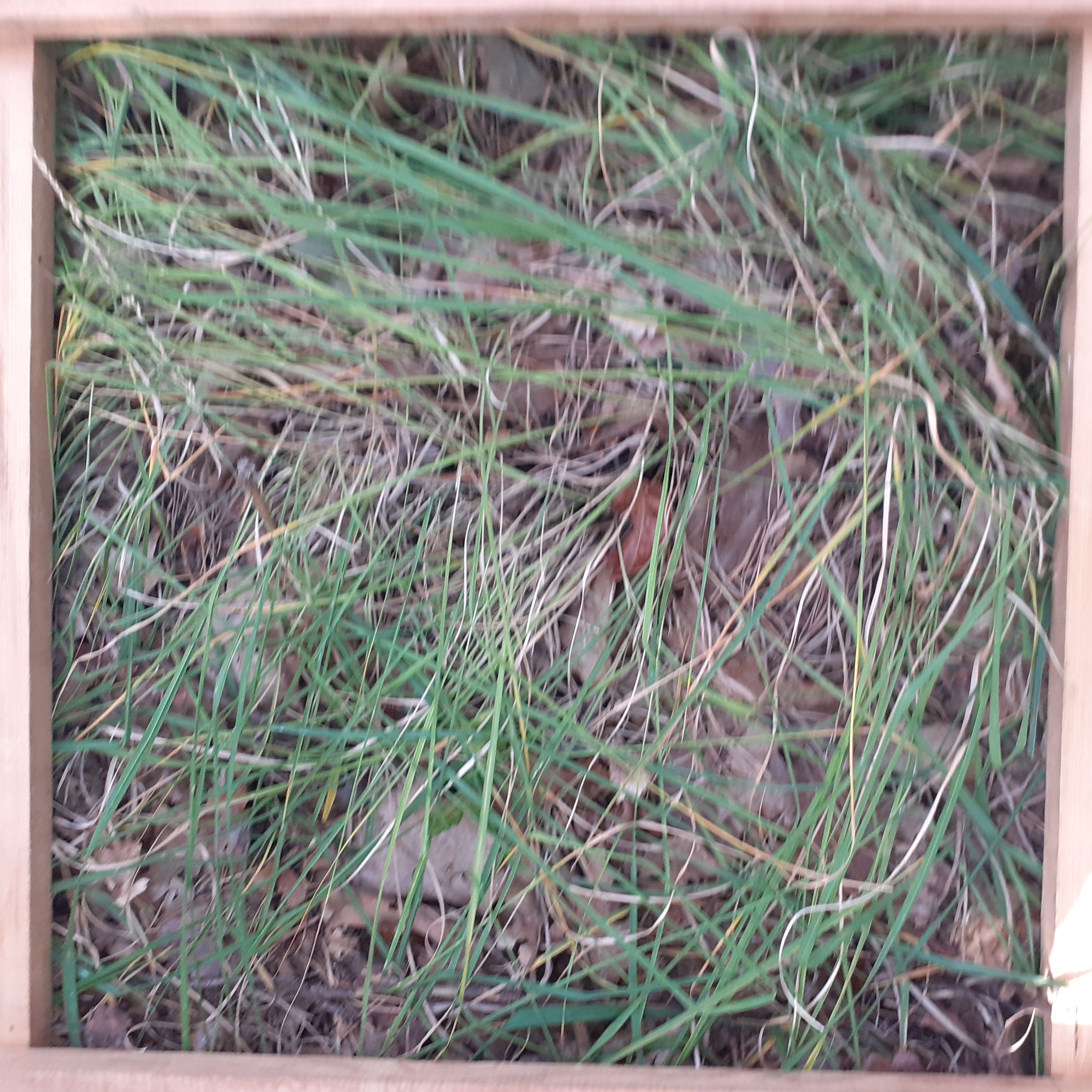} \\
        \includegraphics[width=\linewidth,height=\linewidth]{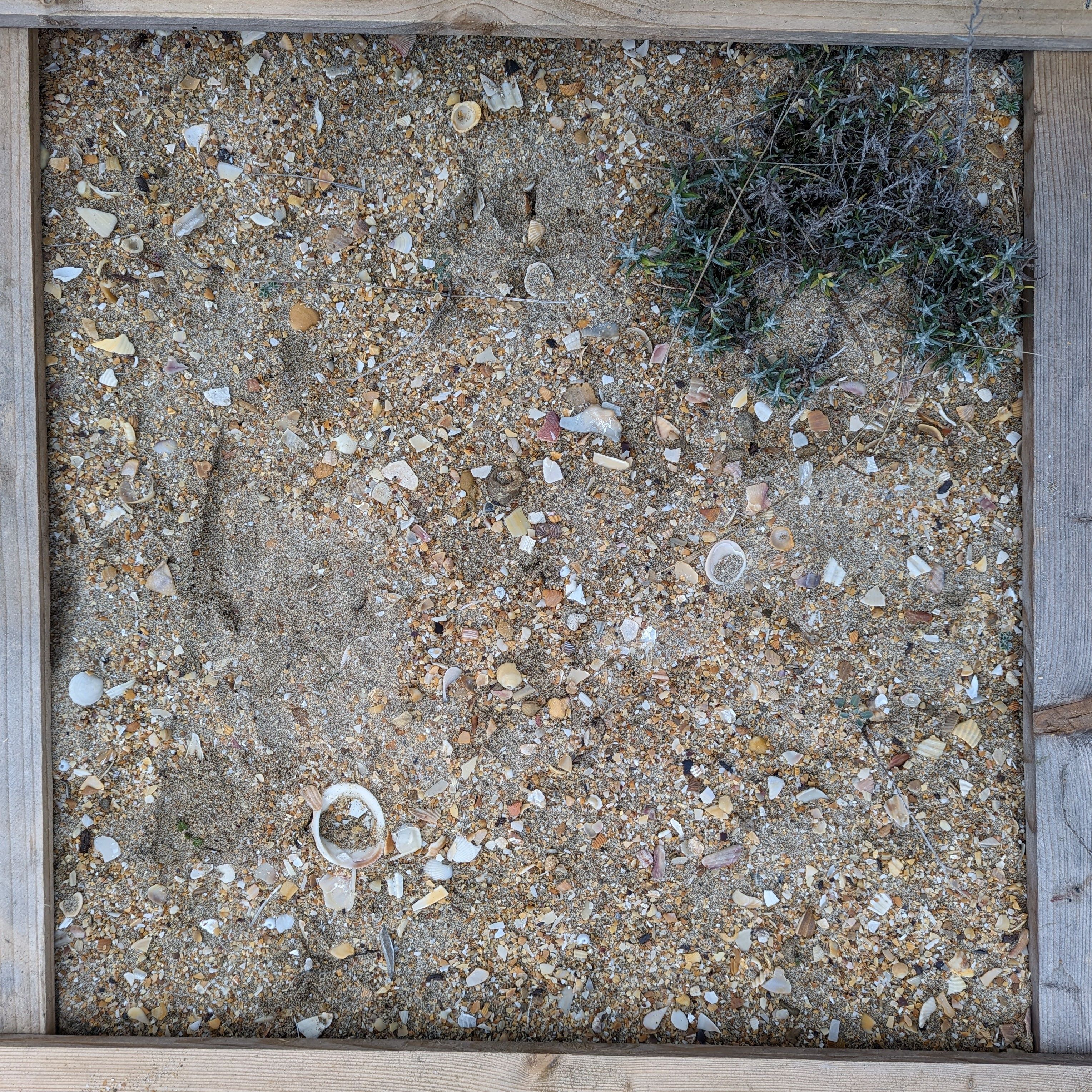} &
        \includegraphics[width=\linewidth,height=\linewidth]{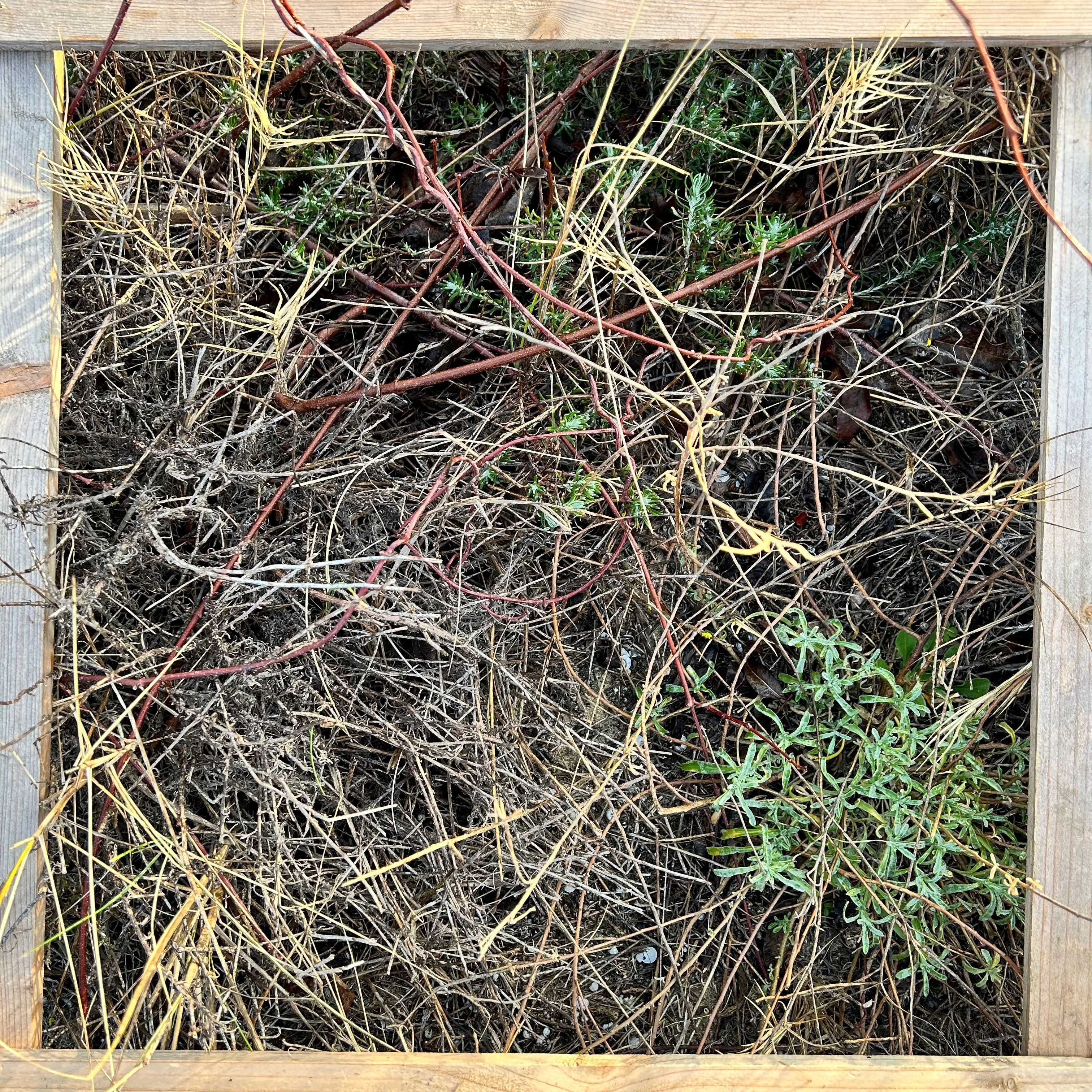} &
        \includegraphics[width=\linewidth,height=\linewidth]{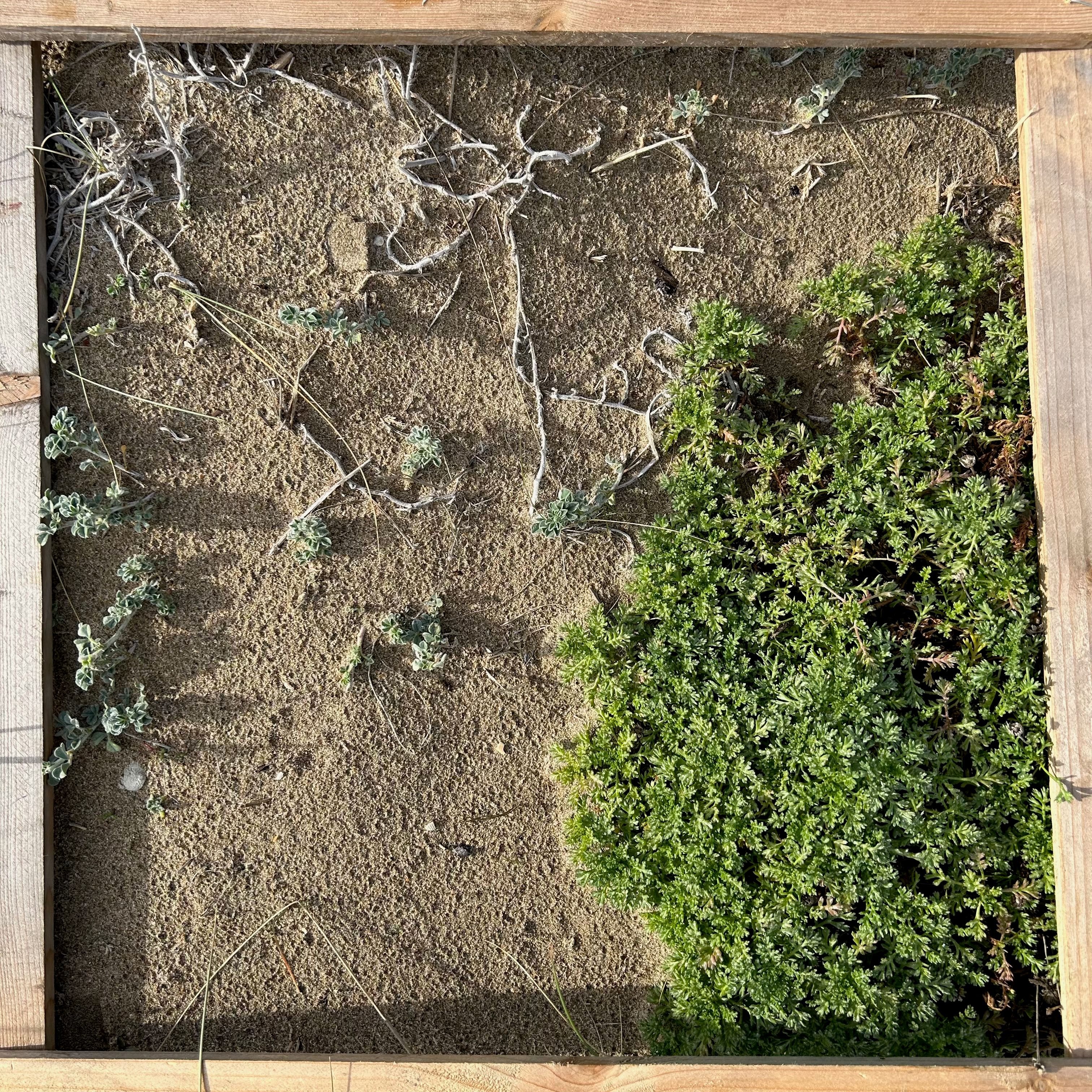} \\
    \end{tabular}    
    \caption{Illustration of the diversity of plot images in the test set.}
    \label{fig:morequadrats}
\end{figure}

The test set is a compilation of several image datasets of quadrats in different floristic contexts, including Pyrenean and Mediterranean floras. These datasets are all produced by experts and consist of a total of 1,695 high-resolution images. The shooting protocol can vary significantly from one context to another: the use of wooden frames or measuring tape to delimit the plot or not, angles of view more or less perpendicular to the ground due to the slope of the site photographed (see Figure \ref{fig:morequadrats}). Additionally, the quality of the images may vary depending on the weather, which can result in more or less pronounced shadows, blurry areas, etc.

\subsection{Pre-trained models}
For participants who may have difficulty finding the computational power necessary to train a plant image identification model on such a large volume of data, or to enable direct work on top of a backbone model for image embedding extraction for instance, two pre-trained models are shared through Zenodo \cite{plantclef2024pretrainedmodel}. Both are based on a state-of-the-art Vision Transformer (ViT) architecture initially pretrained with the DinoV2 Self-Supervised Learning (SSL) approach \cite{oquab2023dinov2, darcet2024vision} and fine-tuned on PlantCLEF 2024 training data.

The vit\_base\_patch14\_reg4\_dinov2\_lvd142m\_onlyclassifier model relies on the original SSL pre-trained weights from the LVD-142M dataset using the DinoV2 SSL method, where only the classification head was trained in a conventional supervised manner, with the backbone being frozen. The second model named vit\_base\_patch14\_reg4\_dinov2\_lvd142m\_onlyclassifier\_then\_all adjusted the backbone weights in a classical supervised manner with the classifier head, resulting in a backbone that no longer aligns with an SSL pre-trained model. 

The models have been pre-trained via the timm library \cite{rw2019timm} and if we follow the way models are named in this library the second model vit\_base\_patch14\_reg4\_dinov2\_lvd142m\_onlyclassifier\_then\_all could be named vit\_base\_patch14\_reg4\_dinov2\_lvd142m\_ft\_pc24, expressing the fact that the initial DinoV2 ViT pre-trained model is subsequently finetuned (ft) on the PlantCLEF2024 dataset in a usual supervised way (all layers are fine-tuned). The vit\_base\_patch14\_reg4\_dinov2\_lvd142m corresponds to the backbone only, without the head, meaning the original pre-trained model with DinoV2 on the huge generalist LVD142M dataset of 142 million images. But for ease of reading, we will subsequently refer to these two models as ViTD2PC24OC and ViTD2PC24All for vit\_base\_patch14\_reg4\_dinov2\_lvd142m\_onlyclassifier and vit\_base\_patch14\_reg4\_dinov2\_lvd142m\_onlyclassifier\_then\_all, respectively. ViTD2 will stand for the original pre-trained backbone vit\_base\_patch14\_dinov2\_lvd142m but without the four registers features.

The training set was split into three sub-directories, subdividing the data into (sub) training, validation, and test sets (see Table \ref{tab:stats} for the statistics) to facilitate the training of individual plant identification models, for the organizers and the participants. It is important not to confuse this test set dedicated to individual plant identification with the challenge test set, which contains large multi-species images.

The two pre-trained models were both trained on an A100 octo-GPU node server using the timm library (version 0.9.16) under torch (version 2.2.1+cu121). The ViTD2PC24OC model, with only the classifier head being fine-tuned, was trained for approximately 17 hours over 92 epochs with a batch size of 1280 per GPU and a high learning rate of 0.01 (hyperparameters for data augmentation techniques and other features are detailed in the Zenodo package \cite{plantclef2024pretrainedmodel}). The second model, ViTD2PC24All, was initialized with the weights of the previously trained ViTD2PC24OC model and was then fully fine-tuned (all layers) for approximately 36 hours over 92 epochs with a significantly lower batch size per GPU (144) and learning rate (0.00008). Both models were trained with a cross-entropy loss since the objective of the training is to predict a single species per image.

Table \ref{tab:pretrained_perf} shows the performances of the two pretrained models on the subset excluded from the training and used as a "Test" set for evaluating performances and model's generalization capacity to identify on individual plants (note again that this "Test" set of individual plants should not be confused with the challenge test set of vegetation plot images).
\begin{table}
    \caption{PlantCLEF 2024 pretrained model top1 and top5 accuracies at the image level.}
    \centering
        \begin{tabular}{llcc}
            \toprule
            Pre-trained model & Short name & Top1 & Top5 \\
            \midrule
            vit\_base\_patch14\_reg4\_dinov2\_lvd142m\_onlyclassifier               & ViTD2PC24OC & 63.69 & 83.88 \\
            vit\_base\_patch14\_reg4\_dinov2\_lvd142m\_onlyclassifier\_then\_all    & ViTD2PC24All& 75.91 & 92.82 \\
            \bottomrule
        \end{tabular}
    \label{tab:pretrained_perf}
\end{table}

 
\section{Task description}
The aim of the challenge is to detect the presence of every plant species on each high-resolution vegetation plot image, from among more than 7,800 species, bearing in mind that plots are generally 50x50cm in size, and that it's rare for there to be dozens and dozens of species simultaneously. Formally, this is a weakly-supervised multi-label classification task for which we only have single-label training data and a significant distribution shift.

\subsection{Metric}

The metric chosen to evaluate the methods of the participants was the F1 score type adapted to finding a good compromise between recall and precision, i.e. not proposing too many species at the risk of being imprecise, but at the same time not proposing too few species at the risk of being incomplete. Among the several variations of F1 score like the macro averaged per class or the micro averaged, the macro-averaged per sample was selected, i.e. the average of the F1 scores calculated individually for each vegetation plot. The other two F1 scores were shown to the participants during the challenge for information purposes, although the macro average calculated per class is no so relevant since not all species are present in the test set, and the micro average is well known to suffer from data imbalances.
More formally, the macro-averaged per sample F1 score is: 
\begin{equation}
\text{F1}_{\text{macro-averaged per sample}} = \frac{1}{N} \sum_{i=1}^{N} F1_i
\end{equation}
where:
\begin{itemize}
    \item $N$ is the number of images in the test set.
    \item $F1_i$ is the F1 score for test image $i$, calculated as:
    \begin{equation}
    F1_i = \frac{2 \cdot \text{Precision}_i \cdot \text{Recall}_i}{\text{Precision}_i + \text{Recall}_i}
    \end{equation}
    \item $\text{Precision}_i$ and $\text{Recall}_i$ are the precision and recall for test image $i$, defined as:
    \begin{align*}
    \text{Precision}_i &= \frac{\text{TP}_i}{\text{TP}_i + \text{FP}_i} \\
    \text{Recall}_i &= \frac{\text{TP}_i}{\text{TP}_i + \text{FN}_i}
    \end{align*}
    \item where for test image $i$:
    \begin{itemize}
        \item $\text{TP}_i$ is the true positives (the number of plant species correctly predicted).
        \item $\text{FP}_i$ is the false positives (the number of plant species incorrectly predicted).
        \item $\text{FN}_i$ is the false negatives (the number of plant species missed ).
    \end{itemize}
\end{itemize}


\subsection{Rules about metadata and additional data}

The use of the metadata (licenses, exif) was authorised provided that, for each run using metadata, an equivalent run using only the visual information without metadata in submitted in order to assess the raw contribution of a purely visual analysis. 

The use of additional data is permitted provided that an equivalent run with only the data provided is submitted to enable more accurate and fair comparisons.

\subsection{Challenge platform}

The PlantCLEF2024 challenge was hosted on the \href{https://huggingface.co/spaces/BVRA/PlantCLEF2024}{Hugging Face platform}, providing an opportunity for researchers and enthusiasts to contribute to the development of plant recognition in such original context of plot analysis.

\section{Participants and methods}
Of the 83 teams officially registered on the CLEF registration system for \href{https://clef2024-labs-registration.dei.unipd.it/}{LifeCLEF}, 34 registered specifically for the PlantCLEF 2024 challenge. On the Hugging Face platform hosting the challenge, 9 teams attempted to submit runs, and in the end 7 teams were able to submit a total of 181 runs (see Table \ref{tab:submissions_number}). Three teams produced and shared working notes, Atlantic \cite{foy2024utilizing}, DS@GT-LifeCLEF \cite{gustineli2024transfer} and NEUON AI \cite{chulif2024patch}, all detailed briefly below. 

Before going into the differences between the approaches of the 3 teams who shared a working note, it's important to note that all have developed a tiling inference pipeline making predictions on sub-images, or "tiles", of a high resolution test image. Indeed, as the training set focuses on individual plant in an image, it is quite intuitive to sub-sample a high-resolution test image in the hope of obtaining tiles focusing each more on a single species. Most have used the pre-trained ViTD2PC24All model, but DS@GT-LifeCLEF and NEUON AI teams have also trained other models that can also be used in their own tiling inference pipeline.

Subsequently, the tiling inference pipelines developed by the participants differed in terms of the initial resolution of the test images and some geometrical parameters to control tiles extraction: tile size, step size, and occasionally a border offset in pixels, which could be useful for excluding the potential presence of wooden frames or measuring tapes in many test images.

Finally, since each tile contributes its set of species predictions, various approaches have been developed to limit (top k), threshold (probabilities), and combine all predictions at the quadrat image scale.

The following subsections describe in more detail what distinguishes the approaches.

\begin{table}
    \caption{Number of submissions by team and shared working notes}
    \centering
    \begin{tabular}{cccc}
        \toprule
        \textbf{Index} & \textbf{Team} & \textbf{Number of submitted runs} & \textbf{Working notes}\\
        \midrule
        0 & Atlantic & 85 & yes\\
        1 & DS@GT-LifeCLEF & 16 & yes\\
        2 & MultiMedia & 28 & no\\
        3 & NEUON AI & 13 & yes\\
        4 & SCaLAR & 3 & no\\
        5 & TerraMapper & 15 & no\\
        6 & yao87 & 21 & no\\
        \bottomrule
        \end{tabular}
\label{tab:submissions_number}
\end{table}

\subsection{Atlantic, Atlantic Technological University \& Technological University Dublin, Ireland, 85 runs \cite{foy2024utilizing}}

This team has invested the most time and number of runs over the 2-month competition period (see Table \ref{tab:submissions_number}), enabling them to gradually gain a good knowledge of the test set data and to better understand and optimize the tile prediction process. 

Like most other teams, their tested approaches rely on the use of the second pre-trained model ViTD2PC24All provided by the challenge organizers and integrated into a tiling inference pipeline. Observing that the resolution of the test images was of the order of 3000 pixels per side, that the plants present could be very small, and that the model took relatively large image sizes (518x518 pixels) as input, this team preferred to extract relatively small 224x224 pixel tiles, which were then upscaled to 518x518 pixels for inference. Several combinations of geometric parameters were tested: with successive tiles overlapping by a third or a quarter, with or without border offset, with or without +/- 90 degree rotation. Subsequently, the 5 species with the highest probabilities are retained for each tile, and a region proposal process groups neighboring tiles that received the same species into potentially larger bounding boxes.

An important observation made by these participants is that the pre-trained model ViTD2PC24All lacks the ability to reject parts of the test images that do not show plants, such as stones, sand, various types of soil, and manufactured objects like measuring tapes, wooden frames, camera bags, pens, notebooks, and shoes. Indeed, this type of data is not present in the training set, and it turns out that high confidence scores occurred on a high percentage of non-plant parts of the test images. To significantly reduce the number of false positives, they combined the powerful Segment-Anything Model (SAM) \cite{kirillov2023segment} with color analysis and geometric rules applied to detected masks, excluding non-plant parts of the test images from the inferences.

This teams also attempted to group predictions from test images taken from the same plot but observed multiple times over time (sometimes over several years). They mentioned that their highest score was achieved by combining three tiling configurations, utilizing SAM for false positive removal, and aggregating predictions from images of the same plot taken over time.

The team made their code available \href{https://github.com/stevefoy/AtlanticAnalytica}{here}.


\subsection{DS@GT-LifeCLEF, Georgia Institute of Technology, USA, 16 runs, \cite{gustineli2024transfer}}.
This team explored different ways of embedding the individual images from the training set to train and compare several classifiers. They used the ViTD2 model, the Vision Transformer pre-trained with DinoV2 on the large generic dataset LVD142M \cite{oquab2023dinov2} in its base version, and the fine-tuned model provided by the organizers, ViTD2PC24All. They first drastically reduced the training dataset by lowering the image resolution to 128x128 pixels while applying a center crop to accelerate the embedding extraction process and the training of the models in general.

The first embedding method is basically and directly the cls\_token, typically used for classification, extracted with the original ViTD2 model. The second embedding method involves taking all outputs of the ViTD2 model for each image, including both the cls\_token and all other tokens, resulting in an output shape of 257x768, which is then reduced to a 64-dimensional vector using a Discrete Cosine Transform (DCT). The third and last embedding method is directly the cls\_token from each image provided by the fine-tuned ViTD2PC24All model.

They trained then three classifiers, three linear layers, one for each embedding dataset, using the Negative Log-Likelihood (NLL) loss. However, this loss, while very similar to the usual cross-entropy, is not specifically designed for multi-label classification; therefore, they used a tiling-based method to address this issue, (the full-image approach in inference was also tested as baselines). They compared 2x2, 3x3, and 5x5 grid approaches, evaluating different top-k species per tile.




\subsection{NEUON AI, NEUON AI \& Swinburne University of Technology Sarawak, Malaysia, 13 runs, \cite{chulif2024patch}} 
This team tested two primary deep learning architectures, Convolutional Neural Networks (CNNs) and Vision Transformers (ViTs), to achieve the best possible results. 
They experimented with a multi-label classification approach using a ViT model, inspired by the RICAP method \cite{Takahashi_2020}. This method involves assembling multiple images of diverse species side by side in various sizes to create large composite images. These new training composites images help to train a models to predict multiple species simultaneously. For that, they replace the head of the provided ViTD2PC24All model by a Multi-Layer Perceptron (MLP) of 3 layers. a sequence of 3 linear layers specifically designed for multi-label classification, assuming that a single layer would not be sufficient to capture the complexity of multiple species prediction combinations to be made.

Like the other teams, a tiling pipeline was used for inference, specifically employing a 4x4 or 8x8 grid approach.

A "top genus" filter was also used to eliminate false positives when several species of the same genus are present within a tile. 

They also incorporated an evaluation of prediction variance within a tile to assess the model’s confidence in its predictions, and potentially eliminate false positives.

Finally, they adapted the Bayesian Model Averaging (BMA) technique to aggregate all scores provided by species predictions for all tiles, as an alternative to pooling or voting.






\section{Results}

\begin{figure}
    \centering
    \includegraphics[width=0.8\linewidth]{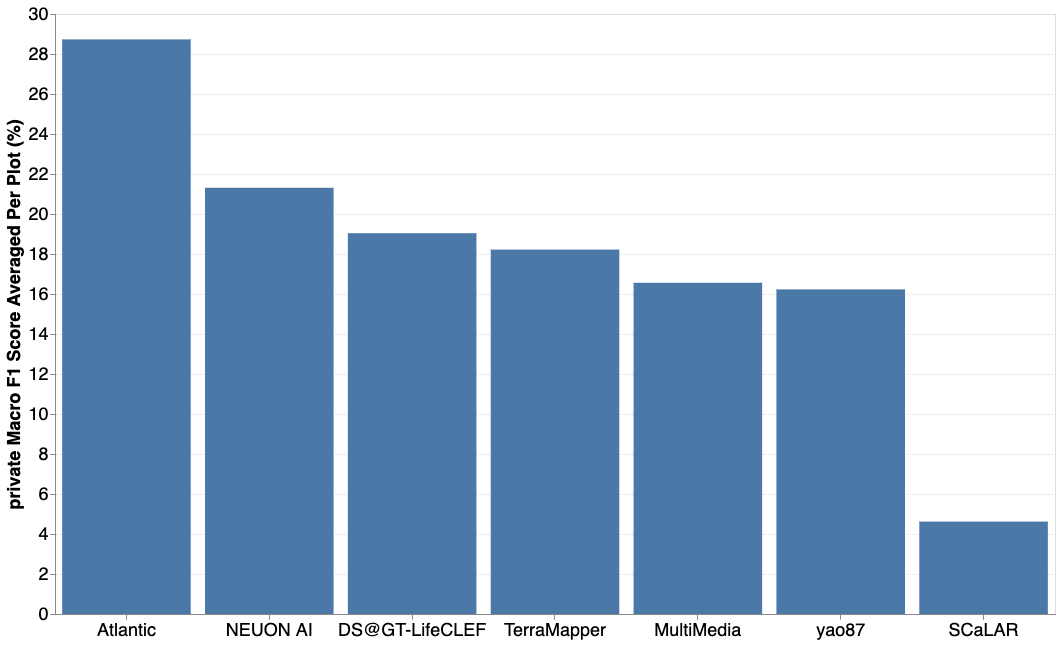}
    \caption{PlantCLEF 2024: top samples F1 scores for each team.}
    \label{fig:PlantCLEF2024Scores}
\end{figure}

Figure~\ref{fig:PlantCLEF2024Scores} reports the best performance achieved for each team as the best private macro-averaged F1 score per sample. The best score, which is clearly ahead of the other methods, was obtained by the Atlantic team with a value of 28,73\%. It was achieved by combining three tiling configurations for inference, utilizing Segment-Anything (SAM) for false positive removal, and aggregating predictions from images of the same plot taken over time. The best score obtained by the NEUON AI team (21.31\%) was actually produced by an ensemble of the predictions of three of their best tested approaches. Their best single approach is based on the provided ViTD2PC24All pre-trained model with a 8x8 tiling inference and using the Bayesian Model Averaging for combining predictions. Team DS@GT-LifeCLEF achieved at best a score of 19.04\% using the ViTD2PC24All pre-trained model as the backbone, with a new classifier predicting and retaining the top1 species for each tile in a 3x3 grid.

\section{Discussion}

Beyond the best performances per team, we can make the following general observations and attempt to analyze in more detail the methodological approaches that seem to have most contributed to these performances.\\
\\
\textbf{A difficult task}: overall performance is quite low and doesn’t exceed a macro-averaged F1 score per sample of 29\%, even by using pre-trained state-of-the-art models on a large volume of data specifically on the flora studied. This demonstrates the significant negative impact of the shift in distribution between the quadrat images and the images of individual plants.\\
\\
\textbf{The highest purely visual performance}, intrinsically linked to the model’s predictions without the use of metadata or rejection mechanisms, is 19.75\% and was obtained by NEUON AI. This run is based on the ViTD2PC24All pre-trained model, using an 8x8 tile grid, aided by the proposed filtering approach on the confidence of prediction scores with the Bayesian Model Averaging. It seems that the Atlantic team achieved a score of 18.47\% by also relying on the ViTD2PC24All model and optimizing the parameters of their inference pipeline. The Atlantic team would therefore have gained almost 10 points (reaching 28.73\%) by excluding predictions for areas that do not contain plants (rocks, sand, soil, non-natural elements like wooden frames or tape measures) with detection based on SAM \cite{kirillov2023segment} and by combining predictions from plots in the same location monitored over time.\\
\\
\textbf{Rejection}: Modeling the rejection of non-plant areas is therefore critical, as demonstrated by the Atlantic team, and to a lesser extent by the NEUON AI team with their confidence thresholding approach. However, this can result in significant computing time costs, taking more than two days to process the entire test dataset of about 1600 images. Integrating true rejection classes (rocks, sand, soil, non-natural elements) in addition to individual plant species into the training dataset could have avoided the need for an additional network. The pre-trained models could have had better performance from the outset.\\
\\
\textbf{Fake quadra generation and multi-label loss.} The use of a method inspired by RICAP (Random Image Cropping And Patching) to train a multi-label deep learning model on a mono-label dataset (as per \cite{Takahashi_2020}) tried by NEUON AI is an interesting idea and yielded a respectable result of 19.08\%, close to their best run with the ViTD2PC24All model (19.75\%). The generation of these fake quadrat images would benefit from better control (avoid close-ups, realistic species combinations, attenuation of collage edge effects, etc). However, like the DS@GT-LifeCLEF team that also learned new classifiers, these two teams may have missed the opportunity to train the new models with a binary cross-entropy loss or asymmetric loss for multi-label classification (ASL) mentioned by the DS@GT-LifeCLEF, which are better suited to the multi-label classification problem, rather than using cross-entropy or Negative Log-Likelihood (NLL) loss.\\
\\
\textbf{About monitoring a site photographed over several years.} Metadata can reveal plots photographed repeatedly over the years, enabling combined predictions for better accuracy. This approach reflects botanists' method of refining identifications through ongoing photo series analysis. It seems that here the gain was high as shown by the Atlantic team but it's risky. In principle, one might expect a plot photographed multiple times in a year over several years to show the same species, but each species has its own growth rhythm, which can be annual or perennial, and the set of species visible can vary from month to month. Identifying them can be more or less difficult depending on growth stages (seedling, flowering, fruiting, senescence). Lastly, consider two images taken in different years but showing the same species at similar growth stages; the lighting and shooting conditions also pose challenges for accurate predictions. \\
\\
\textbf{Multi-scale tiling}: in a quadrat there may be plants of very different sizes, from tiny seedlings to bushes or young shrubs. Multi-scale tiling to adapt to inference or even learning could be interesting for analyzing plants of different sizes, but no team seems to have explored this avenue.\\



\section{Conclusion}
The results of PlantCLEF 2024 show that the problem of multi-label classification of vegetation quadrat HD images is still challenging. Purely visual performance, achieved without metadata or rejection mechanisms, demonstrated that pre-trained models can perform well but benefit greatly from strategies that exclude non-plant areas and combine predictions from plots over time. Modeling the rejection of non-plant areas is crucial, although it can incur substantial computing costs. Integrating true rejection classes into the training dataset could enhance the performance of pre-trained models from the outset. Innovative approaches, such as using RICAP-inspired methods for multi-label deep learning on mono-label datasets, show promise but require better control over fake quadrat generation and appropriate loss functions for multi-label classification. Additionally, utilizing metadata to identify plots photographed over multiple years can improve prediction accuracy, although this approach carries risks due to variations in species visibility and growth stages. Finally, considering multi-scale tiling could be beneficial for analyzing plants of different sizes within quadrats, an avenue yet to be explored.

Beyond its prospects for improvement, the root of the problem remains mainly the shift in distribution between the quadrat images and the images of individual plants in the training set that has a significant negative impact and it is difficult to improve performance even with advanced tiling and/or segmentation approaches. In order to reduce this shift in the future, one solution could be to use unlabelled images of quadrats which are cheaper to produce in large quantities. This would enable the use of unsupervised or self-supervised methods on this type of data. More generally, it is noteworthy that no self-supervised learning (SSL) method has been experimented in the context of PlantCLEF2024 whereas this could have been potentially beneficial even on the individual plants training data. For the next edition of the challenge, we will try to encourage experimentation with self-supervised methods by sharing new data and pre-trained models.

\begin{acknowledgments}

The research described in this paper was partly funded by the European Commission via the GUARDEN and MAMBO projects, which have received funding from the European Union’s Horizon Europe research and innovation program under grant agreements 101060693 and 101060639. The opinions expressed in this work are those of the authors and are not necessarily those of the GUARDEN or MAMBO partners or the European Commission.

This project was provided with computer and storage resources by GENCI at IDRIS thanks to the grant 2023-AD010113641R1 on the supercomputer Jean Zay's the V100 partition.

\end{acknowledgments}


\begin{thebibliography}{12}
\expandafter\ifx\csname natexlab\endcsname\relax\def\natexlab#1{#1}\fi
\providecommand{\url}[1]{\texttt{#1}}
\providecommand{\href}[2]{#2}
\providecommand{\path}[1]{#1}
\providecommand{\DOIprefix}{doi:}
\providecommand{\ArXivprefix}{arXiv:}
\providecommand{\URLprefix}{URL: }
\providecommand{\Pubmedprefix}{pmid:}
\providecommand{\doi}[1]{\href{http://dx.doi.org/#1}{\path{#1}}}
\providecommand{\Pubmed}[1]{\href{pmid:#1}{\path{#1}}}
\providecommand{\bibinfo}[2]{#2}
\ifx\xfnm\relax \def\xfnm[#1]{\unskip,\space#1}\fi
\bibitem[{Go\"{e}au et~al.(2022)Go\"{e}au, Bonnet, and Joly}]{plantclef2022}
\bibinfo{author}{H.~Go\"{e}au}, \bibinfo{author}{P.~Bonnet}, \bibinfo{author}{A.~Joly},
\newblock \bibinfo{title}{Overview of {PlantCLEF} 2022: Image-based plant identification at global scale},
\newblock in: \bibinfo{booktitle}{Working Notes of CLEF 2022 - Conference and Labs of the Evaluation Forum}, \bibinfo{year}{2022}.
\bibitem[{Go\"{e}au et~al.(2023)Go\"{e}au, Bonnet, and Joly}]{plantclef2023}
\bibinfo{author}{H.~Go\"{e}au}, \bibinfo{author}{P.~Bonnet}, \bibinfo{author}{A.~Joly},
\newblock \bibinfo{title}{Overview of {PlantCLEF} 2023: Image-based plant identification at global scale},
\newblock in: \bibinfo{booktitle}{Working Notes of CLEF 2023 - Conference and Labs of the Evaluation Forum}, \bibinfo{year}{2023}.
\bibitem[{Affouard et~al.(2017)Affouard, Goeau, Bonnet, Lombardo, and Joly}]{affouard2017pl}
\bibinfo{author}{A.~Affouard}, \bibinfo{author}{H.~Goeau}, \bibinfo{author}{P.~Bonnet}, \bibinfo{author}{J.-C. Lombardo}, \bibinfo{author}{A.~Joly},
\newblock \bibinfo{title}{Pl@ntnet app in the era of deep learning},
\newblock in: \bibinfo{booktitle}{5th International Conference on Learning Representations (ICLR 2017), April 24-26 2017, Toulon, France}, \bibinfo{year}{2017}.
\bibitem[{Goëau et~al.(2024)Goëau, Lombardo, Affouard, Espitalier, Bonnet, and Joly}]{plantclef2024pretrainedmodel}
\bibinfo{author}{H.~Goëau}, \bibinfo{author}{J.-C. Lombardo}, \bibinfo{author}{A.~Affouard}, \bibinfo{author}{V.~Espitalier}, \bibinfo{author}{P.~Bonnet}, \bibinfo{author}{A.~Joly}, \bibinfo{title}{{PlantCLEF 2024 pretrained models on the flora of the south western Europe based on a subset of Pl@ntNet collaborative images and a ViT base patch 14 dinoV2}}, \bibinfo{year}{2024}. \URLprefix \url{https://doi.org/10.5281/zenodo.10848263}. \DOIprefix\doi{10.5281/zenodo.10848263}.
\bibitem[{Oquab et~al.(2023)Oquab, Darcet, Moutakanni, Vo, Szafraniec, Khalidov, Fernandez, Haziza, Massa, El-Nouby et~al.}]{oquab2023dinov2}
\bibinfo{author}{M.~Oquab}, \bibinfo{author}{T.~Darcet}, \bibinfo{author}{T.~Moutakanni}, \bibinfo{author}{H.~Vo}, \bibinfo{author}{M.~Szafraniec}, \bibinfo{author}{V.~Khalidov}, \bibinfo{author}{P.~Fernandez}, \bibinfo{author}{D.~Haziza}, \bibinfo{author}{F.~Massa}, \bibinfo{author}{A.~El-Nouby}, et~al.,
\newblock \bibinfo{title}{Dinov2: Learning robust visual features without supervision},
\newblock \bibinfo{journal}{arXiv preprint arXiv:2304.07193}  (\bibinfo{year}{2023}).
\bibitem[{Darcet et~al.(2024)Darcet, Oquab, Mairal, and Bojanowski}]{darcet2024vision}
\bibinfo{author}{T.~Darcet}, \bibinfo{author}{M.~Oquab}, \bibinfo{author}{J.~Mairal}, \bibinfo{author}{P.~Bojanowski}, \bibinfo{title}{Vision transformers need registers}, \bibinfo{year}{2024}. \href{http://arxiv.org/abs/2309.16588}{{\tt arXiv:2309.16588}}.
\bibitem[{Wightman(2019)}]{rw2019timm}
\bibinfo{author}{R.~Wightman}, \bibinfo{title}{Pytorch image models}, \bibinfo{howpublished}{\url{https://github.com/rwightman/pytorch-image-models}}, \bibinfo{year}{2019}. \DOIprefix\doi{10.5281/zenodo.4414861}.
\bibitem[{Foy and McLoughlin(2024)}]{foy2024utilizing}
\bibinfo{author}{S.~Foy}, \bibinfo{author}{S.~McLoughlin},
\newblock \bibinfo{title}{Utilizing dino v2 for domain adaptation in vegetation plot analysis},
\newblock in: \bibinfo{booktitle}{Working Notes of CLEF 2024 - Conference and Labs of the Evaluation Forum}, \bibinfo{year}{2024}.
\bibitem[{Gustineli et~al.(2024)Gustineli, Miyaguchi, and Stalter}]{gustineli2024transfer}
\bibinfo{author}{M.~Gustineli}, \bibinfo{author}{A.~Miyaguchi}, \bibinfo{author}{I.~Stalter},
\newblock \bibinfo{title}{Transfer learning for multi-label plant species classification with self-supervised vision transformers},
\newblock in: \bibinfo{booktitle}{Working Notes of CLEF 2024 - Conference and Labs of the Evaluation Forum}, \bibinfo{year}{2024}.
\bibitem[{Chulif et~al.(2024)Chulif, Ishrat, Chang, and Lee}]{chulif2024patch}
\bibinfo{author}{S.~Chulif}, \bibinfo{author}{H.~A. Ishrat}, \bibinfo{author}{Y.~L. Chang}, \bibinfo{author}{S.~H. Lee},
\newblock \bibinfo{title}{Patch-wise inference using pre-trained vision transformers: Neuon submission to plantclef 2024},
\newblock in: \bibinfo{booktitle}{Working Notes of CLEF 2024 - Conference and Labs of the Evaluation Forum}, \bibinfo{year}{2024}.
\bibitem[{Kirillov et~al.(2023)Kirillov, Mintun, Ravi, Mao, Rolland, Gustafson, Xiao, Whitehead, Berg, Lo et~al.}]{kirillov2023segment}
\bibinfo{author}{A.~Kirillov}, \bibinfo{author}{E.~Mintun}, \bibinfo{author}{N.~Ravi}, \bibinfo{author}{H.~Mao}, \bibinfo{author}{C.~Rolland}, \bibinfo{author}{L.~Gustafson}, \bibinfo{author}{T.~Xiao}, \bibinfo{author}{S.~Whitehead}, \bibinfo{author}{A.~C. Berg}, \bibinfo{author}{W.-Y. Lo}, et~al.,
\newblock \bibinfo{title}{Segment anything},
\newblock in: \bibinfo{booktitle}{Proceedings of the IEEE/CVF International Conference on Computer Vision}, \bibinfo{year}{2023}, pp. \bibinfo{pages}{4015--4026}.
\bibitem[{Takahashi et~al.(2020)Takahashi, Matsubara, and Uehara}]{Takahashi_2020}
\bibinfo{author}{R.~Takahashi}, \bibinfo{author}{T.~Matsubara}, \bibinfo{author}{K.~Uehara},
\newblock \bibinfo{title}{Data augmentation using random image cropping and patching for deep cnns},
\newblock \bibinfo{journal}{IEEE Transactions on Circuits and Systems for Video Technology} \bibinfo{volume}{30} (\bibinfo{year}{2020}) \bibinfo{pages}{2917–2931}. \URLprefix \url{http://dx.doi.org/10.1109/TCSVT.2019.2935128}. \DOIprefix\doi{10.1109/tcsvt.2019.2935128}.

\end{thebibliography}





\end{document}